\pdfoutput=1
\documentclass[11pt]{article}

\usepackage[preprint]{acl}

\usepackage{times}
\usepackage{latexsym}

\usepackage[T1]{fontenc}
\usepackage[utf8]{inputenc}

\usepackage{microtype}

\usepackage{inconsolata}

\usepackage{graphicx}
\usepackage{booktabs}
\usepackage{subcaption}
\usepackage{multirow}
\usepackage{longtable}
\usepackage{array}
\usepackage{caption}
\usepackage{amsmath}
\usepackage{float}
\usepackage{stfloats}
\usepackage{xcolor}
\usepackage{hyperref}

\usepackage{tcolorbox} % Include this in your preamble
\definecolor{lightgray}{rgb}{0.9, 0.9, 0.9} % Lighter gray color
\tcbuselibrary{listingsutf8} % Enable utf-8 support for listings if necessary

\definecolor{darkgreen}{rgb}{0.0, 0.5, 0.0}
\title{Predicting User Intents and Musical Attributes \\ from Music Discovery Conversations}

\author{Daeyong Kwon \\ \texttt{ejmj63@kaist.ac.kr} \\\And
        SeungHeon Doh \\ \texttt{seungheondoh@kaist.ac.kr} \\ 
        Graduate School of Culture Technology, KAIST, South Korea \\\And
        Juhan Nam \\ \texttt{juhan.nam@kaist.ac.kr}}

\begin{document}
\maketitle
\begin{abstract}
Intent classification is a text understanding task that identifies user needs from input text queries. While intent classification has been extensively studied in various domains, it has not received much attention in the music domain. In this paper, we investigate intent classification models for \textit{music discovery conversation}, focusing on pre-trained language models. Rather than only predicting functional needs: \textit{intent classification}, we also include a task for classifying musical needs: \textit{musical attribute classification}. Additionally, we propose a method of concatenating previous chat history with just single-turn user queries in the input text, allowing the model to understand the overall conversation context better. Our proposed model significantly improves the F1 score for both user intent and musical attribute classification, and surpasses the zero-shot and few-shot performance of the pretrained Llama 3 model.
\end{abstract}

\section{Introduction}

\begin{figure}[t]
  \centering
  \includegraphics[width=\columnwidth]{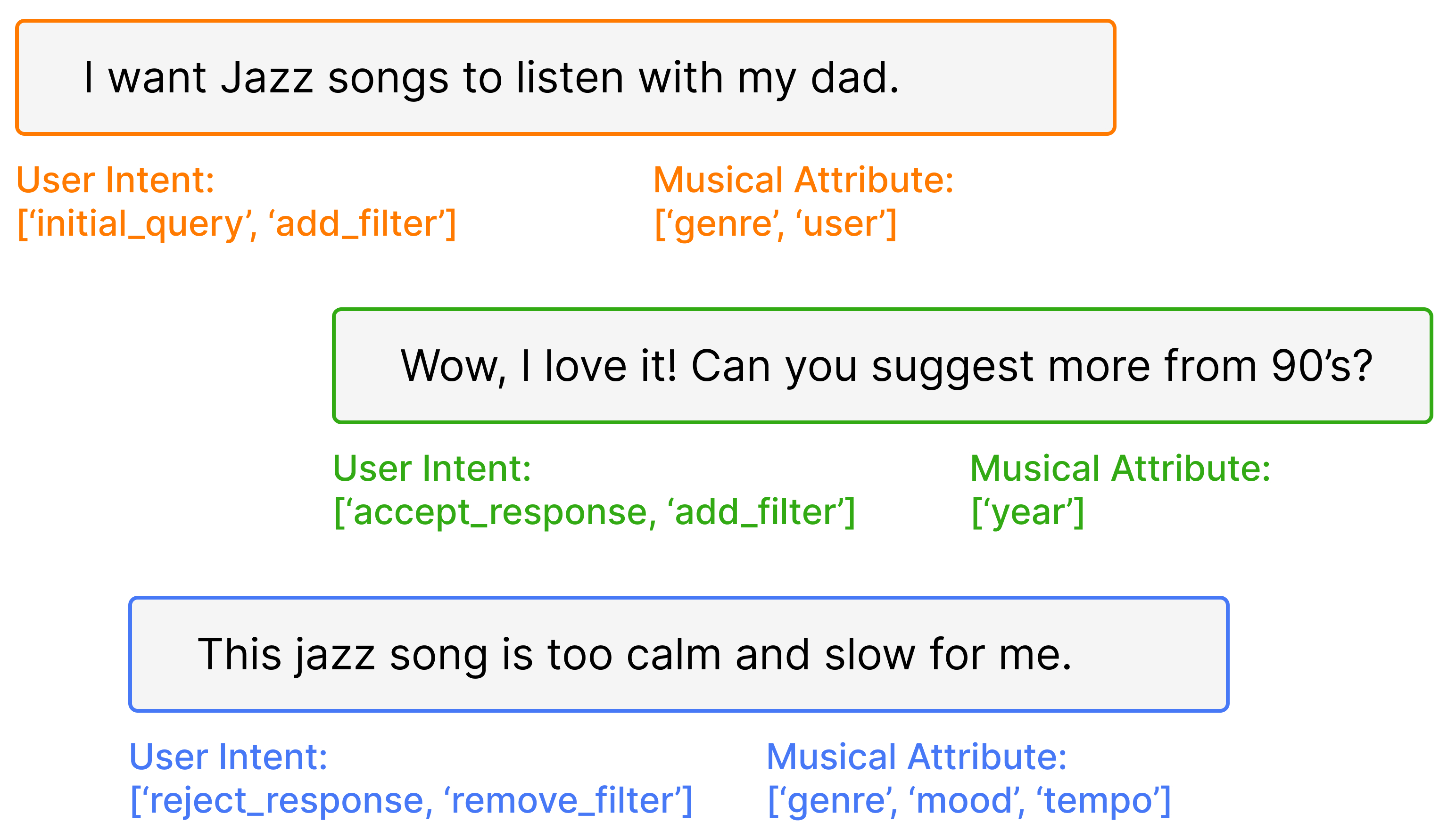}
  \caption{Examples of intent classification for music conversation. Using the given user query as input, the \textbf{user intents} and \textbf{musical attributes} of each conversation are predicted.}
  \label{fig:main_figure}
\end{figure}

Intent classification is a Natural Language Processing (NLP) task that identifies the purpose of user input in conversational systems~\cite{stolcke2000dialogue} and virtual assistants~\cite{weld2022survey}. It determines what the user wants to achieve, enabling the system to respond appropriately and enhance user interaction. Textual user queries are classified into predefined intent categories, typically through the use of discriminative models. For example, in the input \textit{"I want jazz songs to listen to with my dad,"} the system should predict an initial playlist request and apply relevant filters. Additionally, the conversational system should identify that the genre and user are related musical attributes (Figure~\ref{fig:main_figure}).

% SH: review of intent classification methods
The intent classification task has been actively researched alongside advancements in pre-trained language models~\cite{weld2022survey}. Early models focused on task-specific approaches using various features, including sparse representations~\cite{mairesse2009spoken}, word embeddings~\cite{pan2018multiple, bhasin2020parallel}, and BERT-style models~\cite{han2022bi}. More recently, large language models (LLMs)~\cite{touvron2023llama,dubey2024llama}, which are decoder-only transformer models with billions of parameters, have demonstrated strong performance in intent classification through in-context few-shot learning~\cite{loukas2023making}.

% SH: general to domain-specific
Thanks to these advancements in the NLP domain, research on intent classification has been conducted in various fields such as banking~\cite{larson2019evaluation}, travel scheduling~\cite{schuurmans2019intent}, and movie recommendation~\cite{cai2020predicting} to achieve better user query understanding. However, intent classification has received very little attention in the music domain. While research on general domain intent classification~\cite{coucke2018snips} partially covers aspects of musical intent, it only considers two intents (i.e., \textit{PlayMusic} and \textit{AddToPlaylist}) and is limited to single-turn user queries. Accurately identifying the user’s intent and musical needs in music discovery dialogues plays a critical role in enhancing the usability of the chat interface and overall user satisfaction.

% SH: our contirbution
To address this issue, we propose, for the first time, an intent classification task within the context of conversational music retrieval. Our main contributions are as follows: 1) Following prior research~\cite{doh2024lpmusicdialog} that conducted a qualitative analysis of music conversations, we introduce two music-specific intent classification tasks: predicting 8 interface control labels and 15 music attribute labels. 2) We apply and conduct a comparative analysis of various intent classification methodologies from the existing NLP domain. Our findings highlight that current open-source LLMs (e.g., LLaMa3) exhibit more weaknesses than task-specific fine-tuning models in the music conversation domain. 3) Additionally, we identify the optimal use of chat history to improve intent classification performance.

\section{Intent Classification Frameworks}

In our study, intent classification refers to the task of categorizing user input queries into user intent and musical attributes. We used sparse representation, word embedding, DistilBERT, and Llama for the intent classification task.

\subsection{Sparse Representation}

We utilize two types of sparse representations: Bag-of-Words (BoW) and Term Frequency-Inverse Document Frequency (TF-IDF). BoW represents a document by counting word frequencies ignoring grammar and word order, and creates a vector based on a predefined vocabulary. TF-IDF improves on BoW by weighting words based on their frequency in the document relative to their frequency across all documents, giving higher scores to more unique words.

\subsection{Word Embedding}

For dense representation, we used the skip-gram version of Word2Vec~\cite{mikolov2013efficient}, which was trained on the 100B word Google News Corpus and consists of 300-dimensional word vectors for 3M vocabularies. The input sentence is split into words using a whitespace tokenizer, and each word is transformed into a sequence of features through word embedding lookup. The sequence of features is aggregated into a global sentence feature via average pooling.

\subsection{DistilBERT}

Compared to sparse representation and word embedding, BERT~\cite{devlin2018bert} has the advantage of capturing the context of words through bidirectional training, allowing it to understand meaning based on surrounding text.
We use DistilBERT~\cite{sanh2019distilbert}, a smaller version of BERT, to reduce the size of the model and increase its speed. It reduces the size of the model by 40\% while retaining 97\% of its language understanding capabilities and being 60\% faster.

The texts are lowercased and tokenized using the WordPiece tokenizer~\cite{wolf2020transformers}. After tokenization, the token embeddings are processed through 6 transformer blocks to extract 768-dimensional features. For pooling, we use the output from the first position (SOS token) of the feature embeddings as the global sentence feature. It is then utilized for intent classification through MLP layers. We compare a probing model that freezes the DistilBERT and only trains the classifier with a fine-tuning model that adjusts all parameters of the DistilBERT.

\subsection{Llama}
Llama~\cite{touvron2023llama} is a generative model that excels at text generation and performs effectively on large-scale language tasks. We measured the zero-shot and few-shot performance of Llama-3.2 (1B-Instruct and 3B-Instruct) and Llama-3.1 (8B-Instruct)~\cite{dubey2024llama}, which are small-scale models suitable for real-time conversational scenarios, to compare how pretrained general-purpose LLMs perform in the music domain. The input to the Llama model consists of a list of selectable labels and an instruction for classification, and the output is a list of classified labels. In zero-shot tasks, the model makes predictions without any task-specific examples, while few-shot tasks provide 5 examples to guide the model's predictions. Below is an example of references for user intents and musical attributes.

\begin{tcolorbox}[colback=lightgray!30, % Background color
                  colframe=lightgray, % Border color
                  rounded corners, % Rounded corners
                  boxrule=0.5mm, % Thickness of the border
                  width=\linewidth, % Set width to match the column
                  sharp corners=south] % Optional: remove sharp corners at the bottom
    \small{
        \textbf{Input:} \textit{"I want to create a playlist of classical music"} \\
        \textbf{Output:} \texttt{['initial\_query', 'add\_filter']} \\
        \\
        \textbf{Input:} \textit{"Hello, can I get some pop music for a hangout later today?"} \\
        \textbf{Output:} \texttt{['genre', 'theme']}
    }
\end{tcolorbox}

\subsection{Dataset}

We utilized user intent and musical attribute annotations proposed by Doh et al.~\cite{doh2024lpmusicdialog}
For the music discovery conversation taxonomy, Doh et al.~\cite{doh2024lpmusicdialog} employed a grounded theory approach~\cite{khan2014qualitative} to analyze the existing human-to-human music dialogue dataset (CPCD~\cite{chaganty2023beyond}). As a result, they proposed a taxonomy of 8 user intents and 15 musical attributes, with three annotators annotating 888 dialogues in a multi-label format for user intents and musical attributes.

\subsection{Evaluation Metric}
For our multi-label classification task, we used the macro-averaged F1 score to give equal weight to each label, ensuring a comprehensive evaluation across all labels. The best performance threshold for each class label was determined on the validation set (ranging from 0 to 1 with 0.01 increments) and applied to the test set. Values above the threshold were labeled as 1, and those below as 0.

\subsection{Training Setup Details}
The dataset was split into train, validation, and test sets in an 8:1:1 ratio, preserving the proportion of each label. We used the following hyperparameter settings: a batch size of 64, 15 epochs, the Adam optimizer~\cite{kingma2014adam}, and a step learning rate scheduler that decreases the learning rate of the optimizer by a factor of 0.9 at the end of each epoch. The learning rate was set to 2e-4. For the concatenated setting, model was fine-tuned using concatenated sentence inputs.
% The same test dataset was used for evaluating Llama models.
% The specific prompts used for testing can be found in the appendix \ref{app:prompts}.

\begin{table*}[!t]
\fontsize{8}{10}\selectfont
\centering
\begin{tabular}{@{}lcccccccccccc@{}}
\toprule
\multirow{2}{*}{Tag} & \multicolumn{6}{c}{Llama} & \multicolumn{2}{c}{Sparse Representation} & Word Embedding & \multicolumn{2}{c}{DistilBERT} \\ \cmidrule(lr){2-7} \cmidrule(lr){8-9} \cmidrule(lr){10-10} \cmidrule(lr){11-12}
                & 1B\_0 & 1B\_5 & 3B\_0 & 3B\_5 & 8B\_0 & 8B\_5 &  \hspace{5pt} TF-IDF & BoW   & Word2Vec & Probing & Finetune \\ \midrule
Initial Query   & 0.00      & 0.04  & 0.41  & 0.79  & 0.46  & 0.41  &\hspace{5pt} 0.87   & 0.91  & 0.93     & 0.90      & 0.97     \\
Greeting        & 0.10   & 0.17  & 0.32  & 0.73  & 0.72  & 0.62  &\hspace{5pt} 0.89   & 0.96  & 0.80     & 0.95      & 0.99     \\
Add Filter      & 0.85   & 0.72  & 0.49  & 0.80  & 0.75  & 0.91  &\hspace{5pt} 0.93   & 0.92  & 0.90     & 0.94      & 0.96     \\
Remove Filter   & 0.14   & 0.00     & 0.30  & 0.13  & 0.39  & 0.61  &\hspace{5pt} 0.00   & 0.50  & 0.36     & 0.38      & 0.76     \\
Continue        & 0.12   & 0.00     & 0.18  & 0.15  & 0.21  & 0.24  &\hspace{5pt} 0.61   & 0.46  & 0.32     & 0.62      & 0.80     \\
Accept Response & 0.62   & 0.02  & 0.60  & 0.58  & 0.72  & 0.82  &\hspace{5pt} 0.87   & 0.88  & 0.81     & 0.88      & 0.95     \\
Reject Response & 0.08   & 0.00     & 0.00     & 0.17  & 0.36  & 0.62  &\hspace{5pt} 0.17   & 0.36  & 0.50     & 0.65      & 0.82     \\ \midrule
Macro Avg       & 0.27   & 0.14  & 0.33  & 0.48  & 0.52  & 0.61  &\hspace{5pt} 0.62   & 0.71  & 0.66     & 0.76      & 0.89     \\ \bottomrule
\end{tabular}
% \vspace{-0.7em}
\caption{F1 scores for user intents: "0" indicates zero-shot, and "5" refers to few-shot (5-shot) performance.}
\label{tab:f1score_user_intents}
\end{table*}

\begin{table*}[!t]
\fontsize{8}{10}\selectfont
\centering
\begin{tabular}{@{}lcccccccccccc@{}}
\toprule
\multirow{2}{*}{Tag} & \multicolumn{6}{c}{Llama} & \multicolumn{2}{c}{Sparse Representation} & Word Embedding & \multicolumn{2}{c}{DistilBERT} \\ \cmidrule(lr){2-7} \cmidrule(lr){8-9} \cmidrule(lr){10-10} \cmidrule(lr){11-12}
                & 1B\_0 & 1B\_5 & 3B\_0 & 3B\_5 & 8B\_0 & 8B\_5 &  \hspace{5pt} TF-IDF & BoW   & Word2Vec & Probing & Finetune \\ \midrule
Track           & 0.13  & 0.16  & 0.07  & 0.16  & 0.19  & 0.19  &\hspace{5pt} 0.10   & 0.35  & 0.31     & 0.39      & 0.73     \\
Artist          & 0.52  & 0.52  & 0.62  & 0.57  & 0.69  & 0.76  &\hspace{5pt} 0.82   & 0.82  & 0.72     & 0.89      & 0.94     \\
Year            & 0.09  & 0.10  & 0.11  & 0.19  & 0.17  & 0.32  &\hspace{5pt} 0.44   & 0.67  & 0.31     & 0.75      & 0.89     \\
Popularity      & 0.01  & 0.02  & 0.02  & 0.03  & 0.02  & 0.04  &\hspace{5pt} 0.00   & 0.29  & 0.05     & 0.10      & 0.62     \\
Culture         & 0.02  & 0.02  & 0.02  & 0.03  & 0.02  & 0.05  &\hspace{5pt} 0.00   & 0.00  & 0.00     & 0.36      & 0.40     \\
Similar Track   & 0.02  & 0.01  & 0.01  & 0.01  & 0.08  & 0.04  &\hspace{5pt} 0.00   & 0.33  & 0.04     & 0.33      & 0.80     \\
Similar Artist  & 0.05  & 0.05  & 0.05  & 0.10  & 0.20  & 0.15  &\hspace{5pt} 0.14   & 0.34  & 0.25     & 0.52      & 0.74     \\
User            & 0.00  & 0.03  & 0.04  & 0.03  & 0.01  & 0.04  &\hspace{5pt} 0.00   & 0.47  & 0.20     & 0.25      & 0.71     \\
Theme           & 0.15  & 0.18  & 0.19  & 0.30  & 0.23  & 0.35  &\hspace{5pt} 0.70   & 0.76  & 0.63     & 0.74      & 0.89     \\
Mood            & 0.06  & 0.11  & 0.09  & 0.17  & 0.11  & 0.20  &\hspace{5pt} 0.24   & 0.57  & 0.55     & 0.47      & 0.75     \\
Genre           & 0.21  & 0.26  & 0.24  & 0.36  & 0.32  & 0.47  &\hspace{5pt} 0.71   & 0.83  & 0.67     & 0.80      & 0.93     \\
Instrument      & 0.00  & 0.01  & 0.01  & 0.02  & 0.04  & 0.04  &\hspace{5pt} 0.00   & 0.44  & 0.00     & 0.22      & 0.71     \\
Vocal           & 0.00  & 0.02  & 0.02  & 0.04  & 0.02  & 0.05  &\hspace{5pt} 0.00   & 0.40  & 0.13     & 0.25      & 0.35     \\
Tempo           & 0.02  & 0.02  & 0.02  & 0.05  & 0.03  & 0.06  &\hspace{5pt} 0.00   & 0.17  & 0.10     & 0.20      & 0.63     \\ \midrule
Macro Avg       & 0.09  & 0.11  & 0.11  & 0.15  & 0.15  & 0.20  &\hspace{5pt} 0.23   & 0.46  & 0.28     & 0.45      & 0.72     \\ \bottomrule
\end{tabular}
% \vspace{-0.7em}
\caption{F1 scores for musical attributes: "0" indicates zero-shot, and "5" refers to few-shot (5-shot) performance.}
\label{tab:f1score_music_intents}
\end{table*}

\subsection{Concatenate Previous Dialogue History}

The music discovery conversation dataset is characterized as a multi-turn chat dataset between the recommender and the music seeker. Through turn-taking, we can perform intent classification that considers the previous context. In the movie intent classification, Cai et al.~\cite{cai2020predicting} reported that the classification performance of user intent and satisfaction significantly improves by incorporating context features into the classification model. Following previous study~\cite{cai2020predicting}, we also compare the performance of intent classification using concatenated text sentences from previous dialogue turns with the case where only the current query is used.

\section{Result}
\subsection{Performance Comparison}

Tables \ref{tab:f1score_user_intents} and \ref{tab:f1score_music_intents} show the performances of intent classification frameworks. The fine-tuned DistilBERT outperforms all the other models for both user intent and musical attribute classification. While other models had difficulty understanding musical attributes compared to user intents, fine-tuned model handled both tasks effectively. The significant improvement in musical attribute classification (0.46$\rightarrow$0.72) shows that our model has effectively acquired musical knowledge.

Sparse representation, word embedding, and probing models struggled with classifying less frequent labels, such as \textit{remove\_filter}, \textit{continue}, and \textit{reject\_response} for user intent, and \textit{popularity}, \textit{culture}, \textit{similar\_track}, \textit{instrument}, \textit{vocal}, and \textit{tempo} for musical attributes. The fine-tuned model significantly improved performance on these labels, demonstrating that it can achieve high performance even with fine-tuning on a small amount of data.

Also, musical attributes exhibited different results depending on the complexity of the words used. For instance, \textit{popularity} saw a significant performance improvement after fine-tuning (0.10$\xrightarrow{}$0.62) due to the high repitition of words like \textit{'popular'} and \textit{'hits'}. In contrast, \textit{culture} faced more difficulty (0.36$\xrightarrow{}$0.40) as it includes a wide range of words related to different cultures and countries.

The general-purpose Llama models demonstrated lower performance compared to our model fine-tuned on music-domain-specific data, indicating that they lacks sufficient knowledge of the music domain. The Llama models demonstrated improved performance as its size increased; however, using very large models such as 70B and 405B presents challenges in conversational situations that require real-time feedback due to the computational constraints. While Llama-3.1-8B-Instruct model achieved a few-shot performance of 0.61 for user intents, it only showed a few-shot performance of 0.20 for musical attributes, highlighting its difficulty in understanding musical attributes.

\subsection{Concatenating Previous Dialogue History}

\begin{figure}[!t]
  \centering
  \includegraphics[width=0.9\columnwidth]{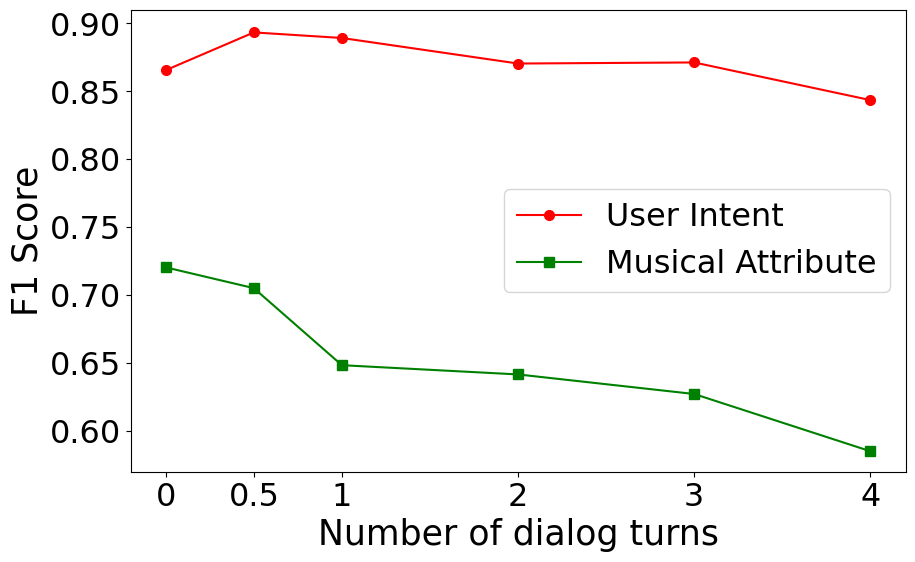}
  % \vspace{-1mm}
  \caption{F1 score comparison by varying context length. The X-axis value 0 represents considering only the current query, 0.5 represents considering the previous query, and 1 to 4 represents the number of considered previous turns.}
  \label{fig:f1score_per_turns}
  % \vspace{-3.5mm}
\end{figure}

Figure \ref{fig:f1score_per_turns} shows the F1 score by varying the consideration of the previous dialogue turns, using fine-tuned model. For user intents, the best performance was achieved when considering only the previous query, which is an X-value of 0.5. Including more context beyond this point decreases performance. User intents exhibit local characteristics, where considering only the previous query is often sufficient. For instance, \textit{initial\_query} and \textit{greeting} can be determined based on the current query alone, while \textit{add\_filter} and \textit{continue} can be inferred by reviewing only the previous query.
% Dialogue turns beyond the most recent one tend to be less relevant to the current conversation and may even obscure critical information.
% As a result, the model considering only the immediate previous sentence became our final best model.

For musical attributes, incorporating context led to worse performance compared to using only the current query (X-value of 0). This is because musical attributes were primarily determined by the presence of musical terms in the current query, regardless of the prior context.

% \subsection{Qualitative Results}

% Table~\ref{tab:cpcd_example} shows two examples of conversational intent classification given a user query. We use the best model, a fine-tuned DistilBERT model, for inference. For instance, our model accurately predicted entire user intents and musical attributes for queries like: \textit{"Hi, I’d like to listen to Mike Oldfield music, particularly from the early stages in his career"}.

% However, our model occasionally failed to accurately identify user intents and musical attributes. For example, it could not detect expected user intents such as \textit{accept\_response} and musical attributes such as \textit{popularity} for the query: \textit{"Perfect for the playlist. Do you have any popular songs from the 90s to add to the playlist?"}.

\section{Conclusion}
We proposed a user intent and musical attribute classification model for music discovery conversations. By fine-tuning a pre-trained language model, our model shows significantly enhanced performance in both user intent and musical attribute classification, especially for labels with a small amount of data. This suggests potential applications in fields where well-annotated large-scale data are not available, such as intent classification tasks in music discovery conversations and the development of conversational music recommendation systems.

We also introduced a method of concatenating the previous context, which helps the model better understand user intents throughout the conversation. The context length is crucial for effectively capturing relevant information, and we found that for user intent, considering only the most recent query provides the best results.
% However, this approach decreases performance for musical attribute classification due to the annotation properties of the dataset. Including too much previous context can obscure important information, leading to a decrease in performance as the number of turns increases. Therefore, choosing an adequate context length that captures relevant information is crucial.

Additionally, we evaluated the zero-shot and few-shot classification performance of the Llama 3 models, which performed lower than the domain-specific fine-tuned DistilBERT model. This suggests that general-purpose LLMs like Llama lack sufficient music domain knowledge, making them less effective for intent classification. The dataset\footnote{\href{https://huggingface.co/datasets/seungheondoh/cpcd-intent}{Dataset Huggingface}}, 
code\footnote{\href{https://github.com/DaeyongKwon98/Intent_Classification}{Github Repository}}, 
and models are publicly available.

\newpage
\section{Limitations}

% This section discusses the limitations and potential future works to address them.

\subsection{Failure Cases}

% We previously identified that the pretrained Llama model lacks sufficient musical knowledge in the context of music discovery conversations.
In this section, we will discuss some representative failure cases, which can serve as reference points for future fine-tuning or further developments.

% \subsubsection{User Intents}

The model usually predicted the correct label but often made unnecessary additional predictions. The concatenated previous query is in blue. The correctly predicted intent is in green, and the incorrectly predicted intent is in red.
\newline
\textbf{Query}: \textit{“\textcolor{blue}{[MUSIC]That's a classic. I have added a few more rock anthems for your choice. Any other song, or artist?} Great hits too! Could we move away from the rock genre? Give me some hits that are classics that everyone knows and would love to hear!”}
\newline \textbf{Ground Truth}: [\textcolor{darkgreen}{'add\_filter'}, \textcolor{darkgreen}{'remove\_filter'}, \textcolor{darkgreen}{'accept\_response'}]
\newline \textbf{Prediction}: [\textcolor{darkgreen}{'add\_filter'}, \textcolor{darkgreen}{'remove\_filter'}, \textcolor{red}{'continue'}, \textcolor{darkgreen}{'accept\_response'}, \textcolor{red}{'reject\_response'}]
\newline \newline
Sometimes, the model incorrectly predicted a label or failed to include a necessary label.
\newline
\textbf{Query}: \textit{“\textcolor{blue}{[MUSIC]These are the results that populated. Nothing directly by Handel. Are these okay?} please can you add a little bit of Katherin jenkins”}
\newline \textbf{Ground Truth}: [\textcolor{darkgreen}{'add\_filter'}, \textcolor{red}{‘accept\_response’}]
\newline \textbf{Prediction}: [\textcolor{darkgreen}{'add\_filter'}, \textcolor{red}{‘continue’}]
% \subsubsection{Musical Attributes}
\newline \newline
For musical attributes, the model also predicted many unnecessary labels, similar to what was observed with user intent.
\newline
\textbf{Query}: \textit{“Is it the Corssroads Album?”}
\newline \textbf{Ground Truth}: [\textcolor{darkgreen}{'track'}]
\newline \textbf{Prediction}: [\textcolor{darkgreen}{'track'}, \textcolor{red}{‘artist’}, \textcolor{red}{‘year’}, \textcolor{red}{‘genre’}, \textcolor{red}{‘instrument’}]
\newline \newline
It also occasionally failed to include the necessary musical attributes, as shown below.
\newline
\textbf{Query}: \textit{“I would like to create a 90s Hip Hop/Rap playlist to listen while I clean my house.”}
\newline \textbf{Ground Truth}: [\textcolor{red}{'year'}, \textcolor{darkgreen}{'genre'}, \textcolor{darkgreen}{'theme'}]
\newline \textbf{Prediction}: [\textcolor{darkgreen}{'genre'}, \textcolor{darkgreen}{'theme'}, \textcolor{red}{‘mood’}, \textcolor{red}{‘instrument’}]

\subsection{Wrong Inference}

% In addition to predicting incorrect labels, there were instances where the model produced words that had not been given.

% \subsubsection{User Intents}

For user intent, 3 wrong inferences occurred across the entire test dataset, and in each case, the predicted intents were present in the query but had not been provided. Below is one example.
\newline
\textbf{Query}: \textit{“can you add some 80 'reggae”}
\newline \textbf{Prediction}: [\textcolor{red}{'reggae'}, ...]

% \subsubsection{Musical Attributes}

For musical attributes, 6 wrong inferences occurred, where the model predicted \textit{gender}, which was not a given. In some cases, there was no content related to gender in the provided query.% Below is one such example.
\newline
\textbf{Query}: \textit{“Best Friend is my favorite”}
\newline \textbf{Prediction}: [\textcolor{red}{'gender'}, ...]

\subsection{Model}

The DistilBERT is not a state-of-the-art model. Better performance could be achieved by fine-tuning larger models (e.g., RoBERTa, Llama) on more advanced computing resources.

\subsection{Dataset}

The CPCD dataset has an imbalanced distribution of user intents and musical attributes. For instance, user intents such as \textit{remove\_filter} and \textit{reject\_response}, as well as musical attributes like \textit{culture}, \textit{popularity}, and \textit{vocal}, are underrepresented in the dataset, making fine-tuning difficult. Therefore, a larger dataset with more balanced labels is needed. It is expected that training on an additional dataset, such as the LP-MusicDialog~\cite{doh2024lpmusicdialog} dataset, could mitigate this issue to some extent. Moreover, using generative LLMs for data augmentation could be a good approach to address the data imbalance.

\subsection{Concatenating Previous Dialogue History}

User intents and musical attributes in our dataset exhibit local characteristics, meaning they require very little previous dialogue history for classification. If a dataset with more organically connected queries within the dialogue were used, future research could explore utilizing dialogue history to improve classification performance.

\section{Ethical Considerations}

While user intent and musical attribute classification tasks themselves may not directly introduce ethical risks, the training data used can contain biases that affect outcomes. If the training data reflects demographic biases, the model may reinforce these biases, leading to unfair or harmful recommendations.
% Additionally, misclassification of user intents based on biased data can result in inappropriate responses or suggestions, impacting user experience.
Therefore, it is crucial to implement ethical oversight throughout the model development process, including regular audits of the training data for biases and considerations of the implications of the model's outputs on different user groups. Continuous monitoring and adjustments are necessary to mitigate potential risks.

% Bibliography entries for the entire Anthology, followed by custom entries
%\bibliography{anthology,custom}
% Custom bibliography entries only
\bibliography{main}

\begin{thebibliography}{21}
\providecommand{\natexlab}[1]{#1}

\bibitem[{Bhasin et~al.(2020)Bhasin, Natarajan, Mathur, and Mangla}]{bhasin2020parallel}
Anmol Bhasin, Bharatram Natarajan, Gaurav Mathur, and Himanshu Mangla. 2020.
\newblock Parallel intent and slot prediction using mlb fusion.
\newblock In \emph{2020 IEEE 14th International Conference on Semantic Computing (ICSC)}.

\bibitem[{Cai and Chen(2020)}]{cai2020predicting}
Wanling Cai and Li~Chen. 2020.
\newblock Predicting user intents and satisfaction with dialogue-based conversational recommendations.
\newblock In \emph{Proceedings of the 28th ACM Conference on User Modeling, Adaptation and Personalization}, pages 33--42.

\bibitem[{Chaganty et~al.(2023)Chaganty, Leszczynski, Zhang, Ganti, Balog, and Radlinski}]{chaganty2023beyond}
Arun~Tejasvi Chaganty, Megan Leszczynski, Shu Zhang, Ravi Ganti, Krisztian Balog, and Filip Radlinski. 2023.
\newblock Beyond single items: Exploring user preferences in item sets with the conversational playlist curation dataset.
\newblock In \emph{Proceedings of the 46th International ACM SIGIR Conference on Research and Development in Information Retrieval}, pages 2754--2764.

\bibitem[{Coucke et~al.(2018)Coucke, Saade, Ball, Bluche, Caulier, Leroy, Doumouro, Gisselbrecht, Caltagirone, Lavril et~al.}]{coucke2018snips}
Alice Coucke, Alaa Saade, Adrien Ball, Th{\'e}odore Bluche, Alexandre Caulier, David Leroy, Cl{\'e}ment Doumouro, Thibault Gisselbrecht, Francesco Caltagirone, Thibaut Lavril, et~al. 2018.
\newblock Snips voice platform: an embedded spoken language understanding system for private-by-design voice interfaces.
\newblock \emph{arXiv preprint arXiv:1805.10190}.

\bibitem[{Devlin et~al.(2018)Devlin, Chang, Lee, and Toutanova}]{devlin2018bert}
Jacob Devlin, Ming-Wei Chang, Kenton Lee, and Kristina Toutanova. 2018.
\newblock Bert: Pre-training of deep bidirectional transformers for language understanding.
\newblock \emph{arXiv preprint arXiv:1810.04805}.

\bibitem[{Doh et~al.(2024)Doh, Choi, Kwon, Kim, and Nam}]{doh2024lpmusicdialog}
Seungheon Doh, Keunwoo Choi, Daeyong Kwon, Taesu Kim, and Juhan Nam. 2024.
\newblock Music discovery dialogue generation using human intent analysis and large language models.
\newblock In \emph{Proceedings of the International Society for Music Information Retrieval Conference (ISMIR)}.

\bibitem[{Dubey et~al.(2024)Dubey, Jauhri, Pandey, Kadian, Al-Dahle, Letman, Mathur, Schelten, Yang, Fan et~al.}]{dubey2024llama}
Abhimanyu Dubey, Abhinav Jauhri, Abhinav Pandey, Abhishek Kadian, Ahmad Al-Dahle, Aiesha Letman, Akhil Mathur, Alan Schelten, Amy Yang, Angela Fan, et~al. 2024.
\newblock The llama 3 herd of models.
\newblock \emph{arXiv preprint arXiv:2407.21783}.

\bibitem[{Han et~al.(2022)Han, Long, Li, Weld, and Poon}]{han2022bi}
Soyeon~Caren Han, Siqu Long, Huichun Li, Henry Weld, and Josiah Poon. 2022.
\newblock Bi-directional joint neural networks for intent classification and slot filling.
\newblock \emph{arXiv preprint arXiv:2202.13079}.

\bibitem[{Khan(2014)}]{khan2014qualitative}
Shahid~N Khan. 2014.
\newblock Qualitative research method: Grounded theory.
\newblock \emph{International journal of business and management}, 9(11):224--233.

\bibitem[{Kingma and Ba(2014)}]{kingma2014adam}
Diederik~P Kingma and Jimmy Ba. 2014.
\newblock Adam: A method for stochastic optimization.
\newblock \emph{3rd International Conference for Learning Representations (ICLR)}.

\bibitem[{Larson et~al.(2019)Larson, Mahendran, Peper, Clarke, Lee, Hill, Kummerfeld, Leach, Laurenzano, Tang et~al.}]{larson2019evaluation}
Stefan Larson, Anish Mahendran, Joseph~J Peper, Christopher Clarke, Andrew Lee, Parker Hill, Jonathan~K Kummerfeld, Kevin Leach, Michael~A Laurenzano, Lingjia Tang, et~al. 2019.
\newblock An evaluation dataset for intent classification and out-of-scope prediction.
\newblock \emph{arXiv preprint arXiv:1909.02027}.

\bibitem[{Loukas et~al.(2023)Loukas, Stogiannidis, Diamantopoulos, Malakasiotis, and Vassos}]{loukas2023making}
Lefteris Loukas, Ilias Stogiannidis, Odysseas Diamantopoulos, Prodromos Malakasiotis, and Stavros Vassos. 2023.
\newblock Making llms worth every penny: Resource-limited text classification in banking.
\newblock In \emph{Proceedings of the Fourth ACM International Conference on AI in Finance}, pages 392--400.

\bibitem[{Mairesse et~al.(2009)Mairesse, Gasic, Jurcicek, Keizer, Thomson, Yu, and Young}]{mairesse2009spoken}
Fran{\c{c}}ois Mairesse, Milica Gasic, Filip Jurcicek, Simon Keizer, Blaise Thomson, Kai Yu, and Steve Young. 2009.
\newblock Spoken language understanding from unaligned data using discriminative classification models.
\newblock In \emph{2009 IEEE International Conference on Acoustics, Speech and Signal Processing}, pages 4749--4752. IEEE.

\bibitem[{Mikolov et~al.(2013)Mikolov, Chen, Corrado, and Dean}]{mikolov2013efficient}
Tomas Mikolov, Kai Chen, Greg Corrado, and Jeffrey Dean. 2013.
\newblock Efficient estimation of word representations in vector space.
\newblock \emph{arXiv preprint arXiv:1301.3781}.

\bibitem[{Pan et~al.(2018)Pan, Zhang, Ren, Hou, Li, Liang, and Liu}]{pan2018multiple}
Lingfeng Pan, Yi~Zhang, Feiliang Ren, Yining Hou, Yan Li, Xiaobo Liang, and Yongkang Liu. 2018.
\newblock A multiple utterances based neural network model for joint intent detection and slot filling.
\newblock In \emph{CCKS Tasks}.

\bibitem[{Sanh et~al.(2019)Sanh, Debut, Chaumond, and Wolf}]{sanh2019distilbert}
Victor Sanh, Lysandre Debut, Julien Chaumond, and Thomas Wolf. 2019.
\newblock Distilbert, a distilled version of bert: smaller, faster, cheaper and lighter.
\newblock \emph{5th Workshop on Energy Efficient Machine Learning and Cognitive Computing - NeurIPS}.

\bibitem[{Schuurmans and Frasincar(2019)}]{schuurmans2019intent}
Jetze Schuurmans and Flavius Frasincar. 2019.
\newblock Intent classification for dialogue utterances.
\newblock \emph{IEEE Intelligent Systems}, 35(1):82--88.

\bibitem[{Stolcke et~al.(2000)Stolcke, Ries, Coccaro, Shriberg, Bates, Jurafsky, Taylor, Martin, Ess-Dykema, and Meteer}]{stolcke2000dialogue}
Andreas Stolcke, Klaus Ries, Noah Coccaro, Elizabeth Shriberg, Rebecca Bates, Daniel Jurafsky, Paul Taylor, Rachel Martin, Carol~Van Ess-Dykema, and Marie Meteer. 2000.
\newblock Dialogue act modeling for automatic tagging and recognition of conversational speech.
\newblock \emph{Computational linguistics}.

\bibitem[{Touvron et~al.(2023)Touvron, Lavril, Izacard, Martinet, Lachaux, Lacroix, Rozi{\`e}re, Goyal, Hambro, Azhar et~al.}]{touvron2023llama}
Hugo Touvron, Thibaut Lavril, Gautier Izacard, Xavier Martinet, Marie-Anne Lachaux, Timoth{\'e}e Lacroix, Baptiste Rozi{\`e}re, Naman Goyal, Eric Hambro, Faisal Azhar, et~al. 2023.
\newblock Llama: Open and efficient foundation language models.
\newblock \emph{arXiv preprint arXiv:2302.13971}.

\bibitem[{Weld et~al.(2022)Weld, Huang, Long, Poon, and Han}]{weld2022survey}
Henry Weld, Xiaoqi Huang, Siqu Long, Josiah Poon, and Soyeon~Caren Han. 2022.
\newblock A survey of joint intent detection and slot filling models in natural language understanding.
\newblock \emph{ACM Computing Surveys}, 55(8):1--38.

\bibitem[{Wolf et~al.(2020)Wolf, Debut, Sanh, Chaumond, Delangue, Moi, Cistac, Rault, Louf, Funtowicz et~al.}]{wolf2020transformers}
Thomas Wolf, Lysandre Debut, Victor Sanh, Julien Chaumond, Clement Delangue, Anthony Moi, Pierric Cistac, Tim Rault, R{\'e}mi Louf, Morgan Funtowicz, et~al. 2020.
\newblock Transformers: State-of-the-art natural language processing.
\newblock In \emph{Proceedings of the 2020 conference on empirical methods in natural language processing: system demonstrations}, pages 38--45.

\end{thebibliography}

\newpage
\appendix

\section{Appendix}
\label{sec:appendix}

\subsection{Data Vocabulary Size}

As seen in Tables \ref{tab:f1score_user_intents} and \ref{tab:f1score_music_intents}, the Bag of Words model outperformed both TF-IDF and word embeddings. Notably, in the context of musical attribute classification, it even surpassed the performance of DistilBERT's probing model. This can likely be attributed to a significant overlap in vocabulary between the training and test datasets.

\begin{table}[h]
\centering
\renewcommand{\arraystretch}{1.2}
\begin{tabular}{cccc}
\hline
Train & Test  & Overlap (Ratio) \\ \hline
4,835 & 1,184 & 908 (76.7\%)       \\ \hline
\end{tabular}
\caption{Vocabulary size of Train and Test datasets, with the ratio of Test vocabulary overlapping with Train.}
\label{tab:vocab_size}
\end{table}

\subsection{Uset Intent, Musical Attribute Frequency}

Overall, the models struggled to classify less frequent labels. To facilitate understanding of the results for each label, we present the average number of each user intent and musical attribute per dialogue.

\begin{figure}[h]
  \centering
  \begin{subfigure}[b]{\columnwidth}
    \centering
    \includegraphics[width=\columnwidth]{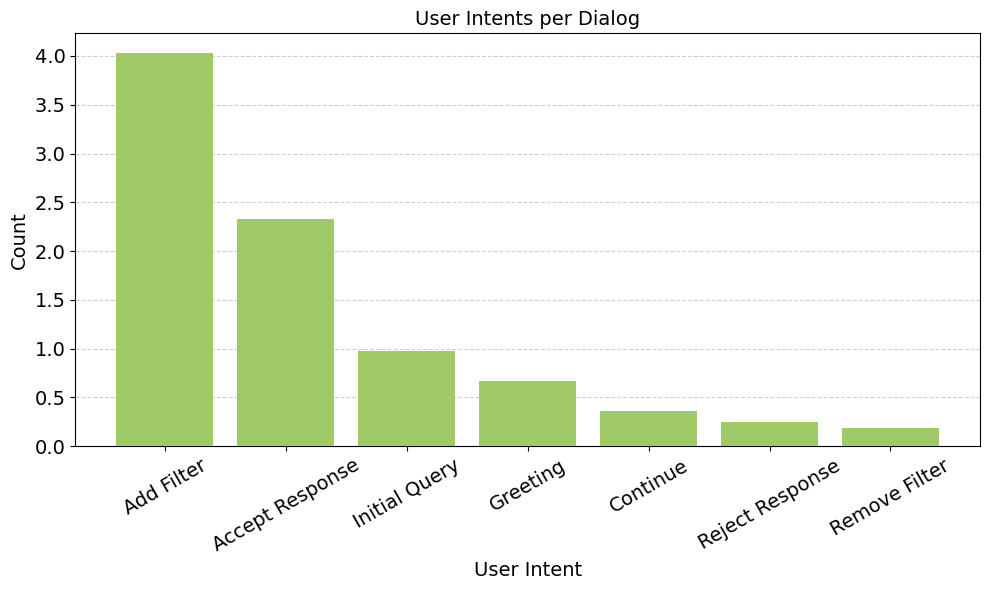}
    \label{fig:user_intents_per_dialog}
  \end{subfigure}
  % \vfill
  \begin{subfigure}[b]{\columnwidth}
    \centering
    \includegraphics[width=\columnwidth]{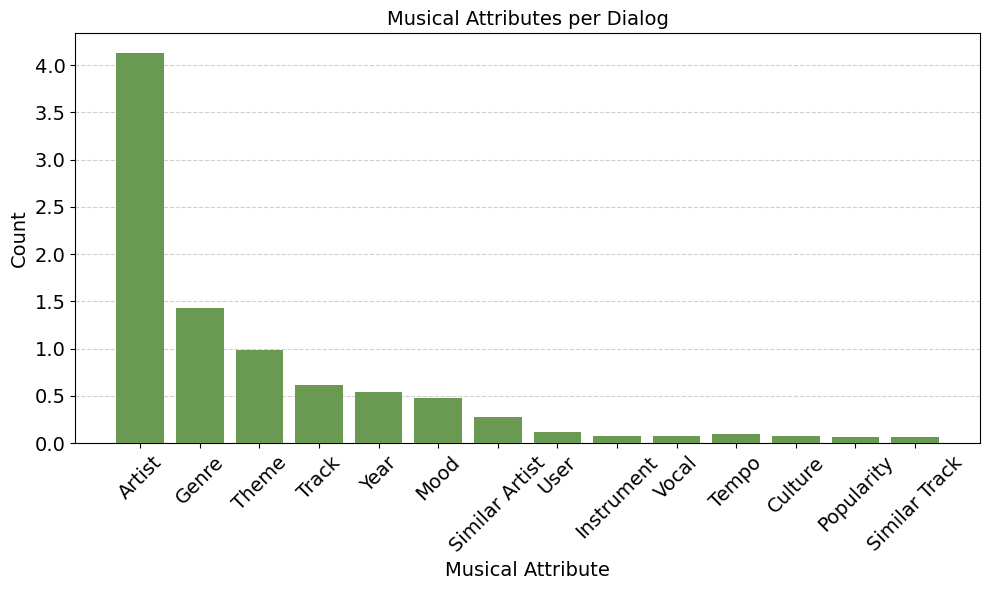}
    \label{fig:musical_attributes_per_dialog}
  \end{subfigure}
  \vspace{-3em}
  \caption{Comparison of Average Counts per Dialog for User Intents and Musical Attributes}
  \label{fig:combined_vertical}
\end{figure}

\subsection{Prompts}
\label{app:prompts}
\subsubsection{Zero-shot Prompt for User Intent}

\textit{"From the following list of user intents: 
[initial\_query, greeting, add\_filter, remove\_filter, continue, accept\_response, reject\_response].
Return only the intents that directly and accurately describe the input text.
Ignore any loosely related or vaguely connected intents.
Provide the result strictly in a list format.
Do not generate any additional text or explanation.
\newline \newline Input: "{input text}"
Output: ["}

\subsubsection{Few-shot Prompt for User Intent}
\textit{"From the following list of user intents: 
[initial\_query, greeting, add\_filter, remove\_filter, continue, accept\_response, reject\_response].
Return only the intents that directly and accurately describe the input text.
Ignore any loosely related or vaguely connected intents.
Provide the result strictly in a list format.
Do not generate any additional text or explanation. \newline
\newline Example: \newline
\newline Input: "I want to listen recent famous songs."
\newline Output: [add\_filter] \newline
\newline Input: "Hello, can you suggest calm music to listen while sleeping?"
\newline Output: [initial\_query, greeting, add\_filter] \newline
\newline Input: "Wow, I love the vide of these songs!"
\newline Output: [accept\_response] \newline
\newline Input: "I think these songs are too fast and loud for me."
\newline Output: [remove\_filter, reject\_response] \newline
\newline Input: "Can you suggest more like these?"
\newline Output: [continue] \newline
\newline Input: "{input text}" Output: ["}

\subsubsection{Zero-shot Prompt for Musical Attribute}

\textit{"From the following list of musical attributes:
[track, artist, year, popularity, culture, similar\_track, similar\_artist, user, theme, mood, genre, instrument, vocal, tempo].
Return only the attributes that directly and accurately describe the input text.
Ignore any loosely related or vaguely connected attributes.
Provide the result strictly in a list format.
Do not generate any additional text or explanation.
\newline \newline Input: "{input text}" Output: ["}

\subsubsection{Few-shot Prompt for Musical Attribute}

\textit{"From the following list of musical attributes:
[track, artist, year, popularity, culture, similar\_track, similar\_artist, user, theme, mood, genre, instrument, vocal, tempo].
Return only the attributes that directly and accurately describe the input text.
Ignore any loosely related or vaguely connected attributes.
Provide the result strictly in a list format.
Do not generate any additional text or explanation.
\newline \newline
Example: \newline
\newline Input: "I want to listen recent famous songs."
\newline Output: [year, popularity] \newline
\newline Input: "Show me faster songs than Ed Sheeran - Shape of You."
\newline Output: [tempo, artist, track] \newline
\newline Input: "Please recommend me some female artists like Rihanna."
\newline Output: [similar\_artist, vocal] \newline
\newline Input: "I need exciting hiphop playlist to listen while I exercise."
\newline Output: [mood, genre, theme] \newline
\newline Input: "African songs to listen with my friends."
\newline Output: [culture, user] \newline
\newline Input: "{input text}" Output: ["}

\end{document}